\documentclass[10pt,twocolumn,letterpaper]{article}

\usepackage{cvpr}              
\usepackage{footmisc}

\usepackage{booktabs}
\usepackage{pgfplots}
\usepackage{multirow}

\usepackage{algorithm}
\usepackage[algo2e]{algorithm2e}

\usepackage{algorithmic}
\usepackage{amsmath}

\usepackage{makecell}

\definecolor{cvprblue}{rgb}{0.21,0.49,0.74}
\usepackage[pagebackref,breaklinks,colorlinks,citecolor=cvprblue]{hyperref}

\def\paperID{4306} 
\def\confName{CVPR}
\def\confYear{2024}

\title{Multi-scale Dynamic and Hierarchical Relationship Modeling for Facial Action Units Recognition}

\author{Zihan Wang$^{1,2,3}$,
Siyang Song$^{4*}$,
Cheng Luo$^{5}$,
Songhe Deng$^{1,2,3}$,
Weicheng Xie$^{1,2,3}$  and 
Linlin Shen$^{1,2,3*}$ \\
	$^1$Computer Vision Institute, School of Computer Science \& Software Engineering, Shenzhen University,\\
        $^2$Shenzhen Institute of Artificial Intelligence and Robotics for Society,\\
	$^3$National Engineering Laboratory for Big Data System Computing Technology, Shenzhen University,\\
        $^4$University of Leicester,      $^5$Monash University\\
}

\begin{document}
\maketitle
\def\thefootnote{$^{*}$}\footnotetext{Corresponding author}
\begin{abstract}



Human facial action units (AUs) are mutually related in a hierarchical manner, as not only they are associated with each other in both spatial and temporal domains but also AUs located in the same/close facial regions show stronger relationships than those of different facial regions. While none of existing approach thoroughly model such hierarchical inter-dependencies among AUs, this paper proposes to comprehensively model multi-scale AU-related dynamic and hierarchical spatio-temporal relationship among AUs for their occurrences recognition. Specifically, we first propose a novel multi-scale temporal differencing network with an adaptive weighting block to explicitly capture facial dynamics across frames at different spatial scales, which specifically considers the heterogeneity of range and magnitude in different AUs' activation. Then, a two-stage strategy is introduced to hierarchically model the relationship among AUs based on their spatial distribution (i.e., local and cross-region AU relationship modelling). Experimental results achieved on BP4D and DISFA show that our approach is the new state-of-the-art in the field of AU occurrence recognition. Our code is publicly available at \url{https://github.com/CVI-SZU/MDHR}.

\end{abstract}
\vspace{-0.3cm}    
\section{Introduction}

\label{sec:intro}
\begin{figure}
    \centering
    \includegraphics[width=0.91\linewidth]{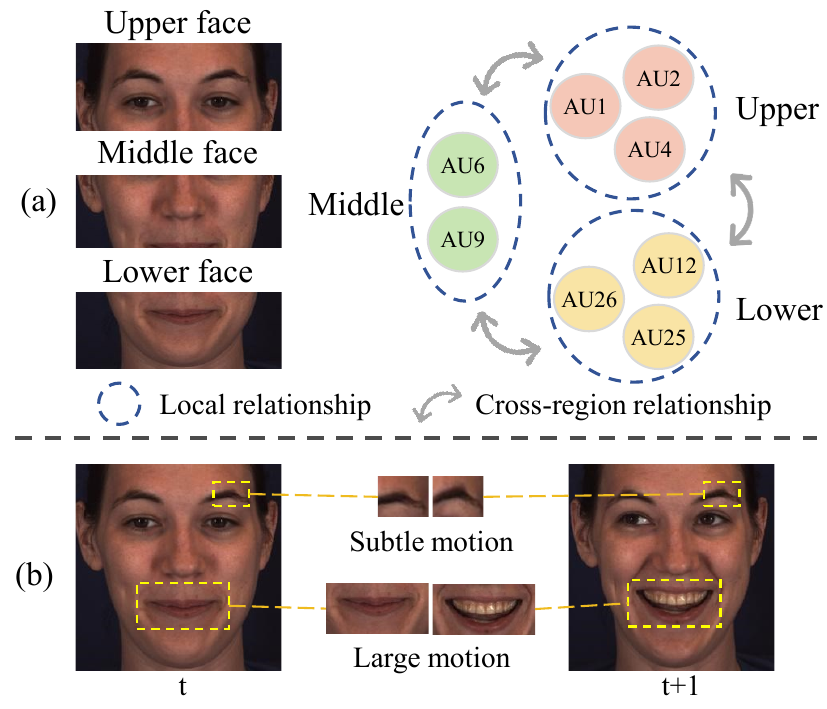}
    \caption{(a) hierarchical AU relationship; and (b) heterogeneous range and magnitude of different AUs' activation.}
    \label{fig:intro}
    \vspace{-0.5cm}
\end{figure}

Facial Action Coding System (FACS) \cite{ekman1978facial} specifies a set of Facial Action Units (AUs) to describe multiple atomic human facial muscle movements, which can comprehensively and objectively describe various human facial expressions in an anonymous and concise manner  \cite{martinez2017automatic}. Recent studies frequently show that AUs are robust and effective low-dimensional facial descriptors for various human behaviours understanding tasks, such as emotion \cite{xie2020assisted,pu2021expression}, mental health \cite{song2020spectral,ringeval2019avec} and pain level \cite{egede2020emopain} analysis. As a result, a large number of studies attempt to automatically recognize AU occurrences from facial images or videos \cite{shao2023facial,song2021uncertain,yang2021exploiting,kaili2016deep,cui2020knowledge,cui2023biomechanics,song2022gratis}.

Most of these approaches conduct AU recognition on still face images. Since each AU usually occur in a specific facial region, some recognize each AU based on a small facial patch defined by automatically detected facial landmarks \cite{7477625, li2017action, shao2021jaa}. However, they not only ignore contextual cues (i.e., AUs are mutually dependent \cite{4293201}) obtainable from other facial regions for each AU's recognition, but also suffer from errors caused by facial landmark detection. Consequently, other approaches jointly recognize multiple AUs from the entire face, allowing informative contextual cues \cite{shao2019facial,niu2019local} to be utilized at the cost of including noises introduced by irrelevant facial regions when recognizing a particular AU. Specifically, transformer \cite{jacob2021facial,yang2023fan} and graph-based \cite{luo2022learning,song2021uncertain} approaches have been widely extended to model the relationship among AUs. However, most of these employ the same strategy to model the relationship between every pair of AUs in the spatial domain(e.g., via transformer encoder with the self-attention operation \cite{jacob2021facial} and graph edges learned by the same cross-attention operation \cite{luo2022learning}), without giving explicit consideration to the natural hierarchical relationship among AUs (\textbf{Problem 1}). More specifically, AUs corresponding to the same/close facial regions frequently show stronger associations than AUs located in different facial regions(illustrated in Fig. \ref{fig:intro} (a)), as AUs localized to the same/close facial regions may be influenced by some shared facial muscles \cite{cattaneo2014facial}.


Besides the spatial cues, some studies additionally model temporal dynamic between facial frames to enhance AU recognition performances \cite{song2022heterogeneous,9413260,li2021integrating}. A typical solution is applying common temporal models (e.g.,  Long-Short-Term-Memory (LSTM) \cite{7961719,li2017action}, Spatio-Temporal Graph \cite{shao2020spatio,wang2023spatio} and 3D Convolution Neural Networks (CNNs) \cite{churamani2021aula}) to process the extracted frame-level static facial/AU features. However, these temporal modeling strategies are insensitive to subtle facial muscle movements. While other approaches \cite{he2022optical,song2019dynamic,yang2023toward} (e.g., optical flow and dynamic image) can explicitly capture facial motions, they still fail to consider that facial muscle movements corresponding to different AUs' activation could exhibit heterogeneity in both range and magnitude (\textbf{Problem 2}), e.g., AU25 involves large-scale deformations of the mouth region, while AU2 are represented by subtle muscle movements surrounding eyebrows (illustrated in Fig. \ref{fig:intro} (b)). In other words, facial dynamic of a certain spatial scale could contribute unequally to the recognition of different AUs.




In this paper, we propose a novel Multi-scale Dynamic and Hierarchical Relationship (MDHR) modeling approach for AU recognition,  which: (i) hierarchically models spatio-temporal relationship among AUs; and (ii) adaptively considers facial dynamic at various spatial scales for each AU's recognition. Our MDHR consists of two key modules. The \textbf{Multi-scale Facial Dynamic Modelling (MFD)} module that adaptively emphasizes AU-related facial dynamic at multiple spatial scales (i.e., computing differences between neighboring frames' features maps output from different backbone layers), ensuring both obvious and subtle AU-related facial dynamic can be captured in an efficient manner (\textbf{addressing Problem 2}). Then, a \textbf{Hierarchical Spatio-temporal AU Relationship Modelling (HSR)} module is introduced to hierarchically model relationship among spatio-temporal AU features in a two-stage manner, where the first stage individually models relationship among AUs within the same/close facial region at both feature extraction and AU prediction levels, and the second stage explicitly learns the relationship between pairs of AUs located in different facial regions via graph edges (\textbf{addressing Problem 1}). The main contributions and novelties of this paper are summarised as follows:
\begin{itemize}

    \item The proposed MFD is the first module that adaptively/specifically considers facial dynamic corresponding to each AU at each spatial scale, as each AUs’ activation exhibit heterogeneity in both range and magnitude.
    
    \item The proposed HSR is the first module that hierarchically learns local and cross-regional spatio-temporal relationship, while previous approaches fail to consider such hierarchical relationship.
    
    \item Experimental results show that our MDHR is the new state-of-the-art on the widely-used AU recognition benchmark datasets: BP4D \cite{zhang2014bp4d} and DISFA \cite{mavadati2013disfa}, where the proposed MFD and HSR modules positively and complementarily contributed to this decent performance.
\end{itemize}

\section{Related Work}
\label{sec:related-work}


\textbf{Static face image-based methods:} Existing approaches frequently predict AUs' status based on static facial displays. Given the anatomical definition of AUs, many of them \cite{7477625,li2017action,li2018eac,corneanu2018deep,jacob2021facial,song2023self} attempted to recognize each AU based on a face patch defined by automatically detected facial landmarks or other prior settings. For example, Zhao et al. \cite{zhao2016deep} proposed a patch-based DRML that learns AU representations robust to variations inherent within local facial regions. EAC-Net \cite{li2018eac} proposed a cropping layer to learn individual AU's representation from small AU-specific areas. Furthermore, JAA-Net \cite{shao2021jaa} jointly conducted AU recognition and face alignment, where the predicted facial landmarks are used to localize each AU region. To take global facial contextual cues into consideration, alternative approaches \cite{shao2019facial,li2021integrating,luo2022learning,shao2022facial,10151797} learn each AU's representation holistically from the full face image, where spatial attention mechanisms have been widely explored. Shao et al. \cite{shao2019facial} employed adaptive channel-wise and spatial attention strategy to enforce the model focusing on AU-related local features from the global face. Li et al. \cite{9666970} proposed a self-diversified multi-channel attention to seek a more robust attention between the global facial representation and each target AU. As AUs are mutually related \cite{Zhang_2018_CVPR}, recent approaches \cite{niu2019local,jacob2021facial,chang2022knowledge,yang2023fan} also specifically modelled the underlying relationship among them. For exampple, LP-Net \cite{niu2019local} applied LSTMs to capture AU relationship. Jacob et al.\cite{jacob2021facial} proposed a transformer-style AU correlation network. In addition, graph-based strategies have been frequently investigated to model AU relationship\cite{luo2022learning,li2019semantic,liu2020relation,song2021uncertain}, where graph nodes have been frequently used to represent target AUs while edges explicitly define the relationship between every pair of AUs.




\textbf{Spatial-temporal methods:} Since facial dynamic also provide crucial cues for AU recognition \cite{chu2017learning}, LSTM has been frequently employed by early studies \cite{jaiswal2016deep,chu2017learning} to model temporal dynamic between static facial features extracted from adjacent frames. To further explore spatio-temporal relationship among AUs, a Spatio-temporal Graph Neural Network (GNN) \cite{shao2020spatio} and a Heterogeneous Spatio-temporal Relation Learning Network (HSTR-Net) \cite{song2022heterogeneous} have been proposed, both of which first construct a set of spatial graphs to model static AU relationship at the frame-level, and then individually model each AU's temporal dynamic by considering its corresponding spatial graph node features across all frames. In addition, Li et al. \cite{Li_2023_ICCV} applied a transformer to learn both spatial AU dependencies and temporal inter-frame contexts by representing the inter-AU and inter-frame correlations within a multi-head attention matrix. Besides such standard temporal model-based feature-level dynamic modelling, other solutions \cite{li2019self,9736614,he2022optical,song2019dynamic,yang2019learning} also have been investigated. For example, two auxiliary AU related tasks (e.g., ROI inpainting and optical flow estimation) are jointly conducted in \cite{yan2022weakly} to enhance the regional features and encode the facial dynamic into the global facial representation, respectively. More recently, Yang et al. \cite{yang2023toward} introduced a temporal difference network (TDN) that extract facial dynamic at a specific spatial scale. Despite the progress made by approaches discussed above, to the best of our knowledge, none of them has specifically modelled AU-related multi-scale facial dynamic and the hierarchical spatio-temporal relationship among AUs.

\section{Methodology}

\begin{figure*}[htbp]
    \centering
    \includegraphics[width=0.96\textwidth]{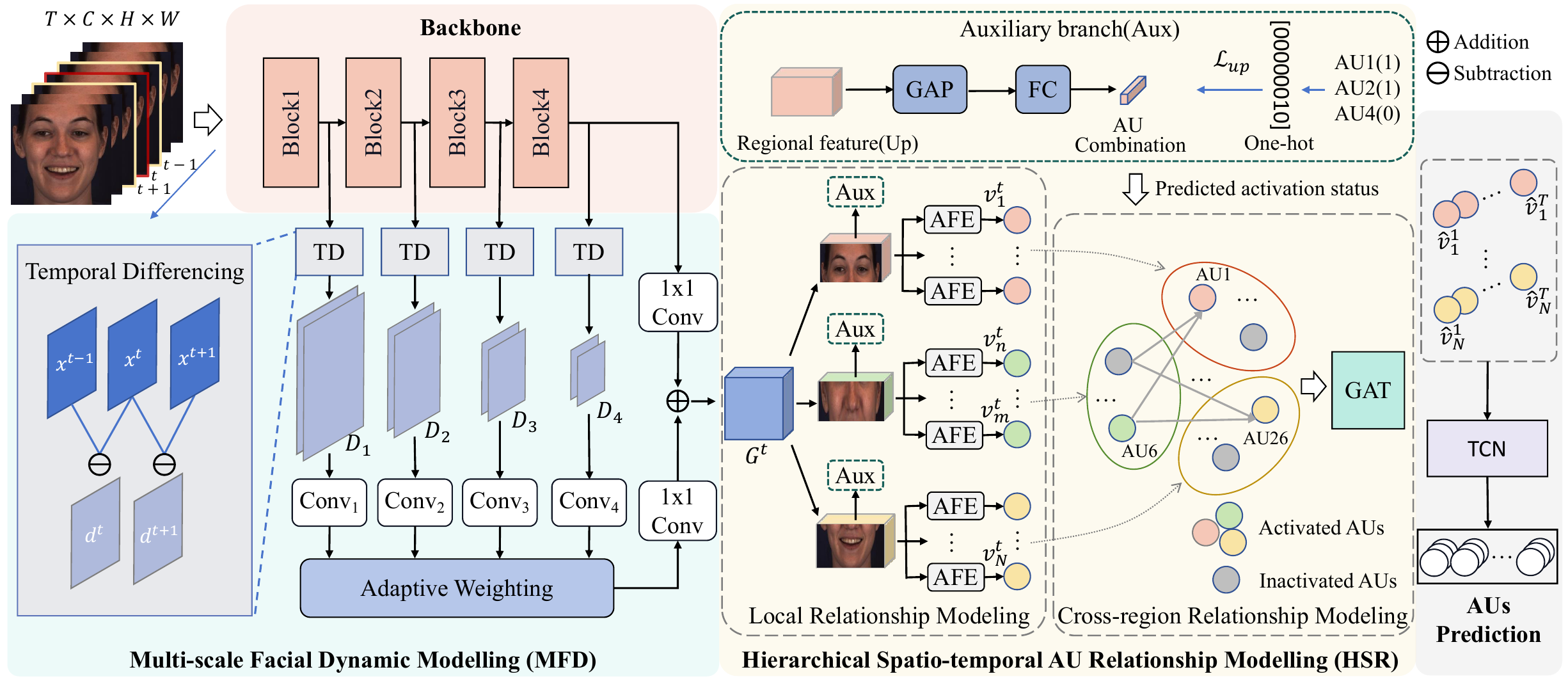}
    \caption{The pipeline of our MDHR, where $k$ is set to 1. The MFD module (Sec. \ref{subsec:AU-dynamic}) first computes facial dynamic at multiple spatial scales based on feature maps output from multiple backbone hidden layers and the output layer. Then, the HSR module (Sec. \ref{subsec:spatio-temporal}) then individually models the relationship among AUs located in the same and different facial regions (the Auxiliary branch is only used at the training phase to make AU combination for each facial region (upper facial region is used as an example in the figure)). Finally, a TCN is individually employed to process every AU feature's sequence of all the input $T$ frames.}
    \label{fig:overview}
    \vspace{-0.3cm}
\end{figure*}

\noindent \textbf{Overview:} Given $T$ consecutive facial frames $S = \{f^1, f^2, \cdots ,f^T\} \in \mathbb{R}^{ T \times C \times H \times W}$ , our approach jointly predicts multiple ($N$) AUs' occurrence at the $t_\text{th}$ facial frame $f^t$ ($t = 1,2,\cdots, T$) by taking not only the $f^t$ but also its adjacent frames into consideration. As illustrated in Fig. \ref{fig:overview} and Algorithm \ref{alg:two}, our MDHR starts with utilizing a backbone (e.g., CNN or Transformer) to jointly extract static facial features from $f_t$ and its adjacent frames $A^t = \{f^{t-k},\cdots,f^{t-1}, f^{t+1},\cdots,f^{t+k}\}$. For each frame, multi-scale static facial features are produced by $L-1$ backbone hidden layers and the output layer (the $L_\text{th}$ layer). Thus, $L$ static facial feature sets corresponding to $2k+1$ frames $X_l = \{x_l^{t-k}, \cdots, x_l^{t}, \cdots, x_l^{t+k} | l = 1,2, \cdots, L\}$ are generated (\textbf{line 2 in Algorithm \ref{alg:two}}). Then, these multi-scale features are fed to the \textbf{Multi-scale Facial Dynamic Modelling (MFD)} module, targeting at not only explicitly capturing facial dynamic at multiple spatial scales, but also adaptively combining these multi-scale facial dynamic features with the static feature $x^t_L$ (\textbf{line 3 in Algorithm \ref{alg:two}}). Based on the spatio-temporal full face representation $G^t$ learned by MFD, a \textbf{Hierarchical Spatio-temporal AU Relationship Modelling (HSR)} module further adaptively models the hierarchical spatio-temporal relationship among AUs in a two-stage manner, where the spatial distribution of the target AUs on the human face is considered, resulting in $N$ individual AU representations $\hat{V}^t = \{\hat{v}^t_1,\cdots,\hat{v}^t_N\}$ (\textbf{line 4 in Algorithm \ref{alg:two}}). Finally, a Temporal Convolution Networks (TCN) \cite{liu2020teinet} with similarity calculating (SC) strategy \cite{luo2022learning} are employed to predict $N$ AUs' occurrences of the input $T$ frames as $P^1, P^2, \cdots, P^T$ ($P^t = \{p^t_1, \cdots, p^t_N \}$, \textbf{line 6 in Algorithm \ref{alg:two}}).

\begin{algorithm}[hbt!]
\caption{Pipeline of the proposed approach (MDHR)}
\label{alg:two}
\SetKwInOut{Input}{Input}\SetKwInOut{Output}{Output}
\Input{$T$ consecutive facial frames $S= \{f^1,\cdots,f^T\}$}
\Output{$N$ AU's predictions of each frame $f^t$}

\begin{algorithmic}[1]
\FOR{$t = 1$ \KwTo $T$}
\STATE Generating multi-scale static global representations $X_1, X_2 \cdots X_L$ $\gets$ $\text{Backbone}(f^{t-k},\cdots,f^{t}\cdots f^{t+k})$ 

\STATE Generating global spatio-temporal features $G^t$ $\gets$ $\text{MFD}(X_1, X_2 \cdots X_L)$

\STATE Generating hierarchical spatio-temporal relationship-aware AU features $\hat{V}^t$ $\gets$ $\text{HSR}(G^t)$
\ENDFOR
\STATE Predicting $N$ AUs of all frames $P^1, P^2, \cdots, P^T$ $\gets$ $\text{SC}(\text{TCN}(\hat{V}^1, \cdots, \hat{V}^t, \cdots, \hat{V}^T))$

\end{algorithmic}
\end{algorithm}
\vspace{-0.2cm}
\subsection{Multi-scale facial dynamic modelling}
\label{subsec:AU-dynamic}

\noindent Inspired by the fact that facial muscle movements are continuous and smooth while each AU exhibit heterogeneity in their range of motions and magnitudes \cite{cattaneo2014facial}, we propose a novel MFD module to model the preceding and proceeding temporal evolution of the target face at multiple spatial scales. It includes a multi-scale Temporal Differecing block that first computes differences between global facial features extracted from every pair of neighboring frames at multiple spatial scales. The obtained multi-scale facial dynamic features are then masked by a set of weighting matrices learned by our adaptive weighting block, aiming to emphasize the informative cues for target AUs at multiple spatio-temporal scales.

\textbf{Multi-scale Temporal Differecing block:} This block is made up of multiple Temporal Differecing (TD) layers followed by convolution layers, which takes feature maps $X_l = \{x_l^{t-k}, \cdots, x_l^{t}, \cdots, x_l^{t+k} | l = 1,2, \cdots, L\}$ produced by multiple ($L-1$) hidden layers and the output layer of the backbone as the input, where $x_l^{t} \in \mathbb{R}^{C_l \times H_l \times W_l}$ denotes the feature map corresponding to the $t_\text{th}$ facial frame generated from the $l_\text{th}$ backbone hidden layer (i.e.,  $C_l$, $H_l$, and $W_l$ represent the channel, height and width of the $x_l^{t}$, respectively). Here, the $l_\text{th}$ TD layer conducts point-to-point subtraction on feature maps produced by the $l_\text{th}$ hidden layer between neighboring frames, aiming to capture facial dynamic at a certain spatial scale. This can be formulated as:
\begin{equation}
    d_l^t = x_l^t-x_l^{t-1}
\end{equation}
Thus, a dynamic feature map $d_l^{t} \in \mathbb{R}^{C_l \times H_l \times W_l}$ representing the facial dynamic between $f^t$ and $f^{t-1}$ at the $l_\text{th}$ spatial scale are produced from the $l_\text{th}$ TD layer. As a result, $L$ sets of dynamic features $D_l = \{d_l^{t-k+1}, \cdots, d_l^{t}, \cdots, d_l^{t+k} | l = 1,2, \cdots, L\}$ are obtained to represent facial dynamic at $L$ different scales. After that, we introduce $L$ step convolution layers to resize the dynamic features extracted at different spatial scales as:
\begin{equation}
    \hat{d}_l^t = \text{Conv2D}_l(d_l^t)
\end{equation}
where the kernel size and stride of the $l_\text{th}$ Conv2D layer $\text{Conv2D}_l$ are set to $8/l$, ensuring all produced dynamic features $\hat{d}_l^t \in \mathbb{R}^{c,h,w}$ to have the same shape. Finally, an average pooling is employed to process all re-shaped dynamic features at each spatial scale along the temporal axis as:
\begin{equation}
    \bar{d}_l^t = \text{Avg}(\hat{d}_l^{t-k+1},\cdots,\hat{d}_l^t,\cdots,\hat{d}_l^{t+k})
\end{equation}
This way, multi-scale and equal-shape facial dynamic features $\bar{d}_1^t, \bar{d}_2^t, \cdots, \bar{d}_L^t$ of the target frame $f^t$ can be obtained, where each $\bar{d}_l^t$ summarizes the temporal evolution of the $f^t$ by considering its preceding and succeeding $k$ frames.


\textbf{Adaptive weighting block:} Facial muscle movements of large range and magnitude are typically associated with feature maps produced from deep backbone layers while subtle facial dynamic usually can be better described by feature maps produced from shallow backbone layers \cite{liu2019learning}. Thus, instead of simply conducting element-wise summation or concatenation (i.e., equally treats all components of all feature maps), we propose to adaptively learn $L$ weighting matrices for properly combining the obtained $L$-scale dynamic features according to the target AUs' typical and unique spatio-temporal scales. In particular, the weight matrix $w_l^t$ at each spatial scale is obtained by exploring the underlying and internal cues from the obtained multi-scale dynamic features, which can be formulated as:
\begin{equation}
    w_l^t = \text{Softmax}(\text{Conv}_l(\text{Concat}([\bar{d_1^t},\bar{d_2^t},\cdots,\bar{d_L^t}]))) 
\end{equation}
where $l = 1, 2, \cdots, L$. Specifically, multi-scale spatio-temporal features $\bar{d_1^t}, \cdots, \bar{d_L^t}$ of the $f_t$ are first concatenated along their channels, followed by $1 \times 1$ convolutions to reduce the number of its channels to one. This results in a unique weighting matrix $w_l^t \in \mathbb{R}^{h \times w}$ to mask the spatio-temporal feature at each spatial scale $l$. A Softmax function is also applied to normalize the obtained weights such that $\sum_{l=1}^{L} w_l^{t,i,j} = 1$ and $w_l^{t,i,j}\in [0,1]$, where $i$ and $j$ index the spatial dimensions. Consequently, each obtained weight matrix $w_l^t$ is applied to the corresponding dynamic feature map $\bar{d}_l^t$ by performing element-wise multiplication as:
\begin{equation}
     x_\text{motion}^t = \sum_{l=1}^{L} w_l^t * \bar{d}_l^t
\end{equation}
where $x_\text{motion}^t$ represents the aggregated and adaptively weighted multi-scale facial dynamic representation of the $f_t$, which is then combined with the spatial feature $x_L^{t}$ produced by the output layer via the element-wise summation:
\begin{equation}
    G^t = x_\text{motion}^t + x_L^{t}
\end{equation}
In summary, the proposed MFD module adaptively incorporates AU-aware facial dynamic with static and global facial cues into $G_t$ for the fine-grained facial AU recognition.








\subsection{Hierarchical spatio-temporal AU relationship modelling}
\label{subsec:spatio-temporal}


\noindent Our HSR module hierarchically models the spatio-temporal relationship among AUs by specifically considering their spatial distribution on the face, as association among AUs in the same/close facial region could be stronger than AUs located in different facial regions \cite{cattaneo2014facial}. It consists of two stages: the \textbf{local AU relationship modelling} stage first models the relationship among AUs located in the same facial region, and then the \textbf{cross-regional AU relationship modeling} stage adaptively explore the relationship between AU pairs of different facial regions.

\textbf{Local AU Relationship Modelling:} This stage specifically models relationship among AUs located in the same facial regions at both their features extraction level and prediction level. It builds on the assumption that constraining each AU feature's extraction to its spatially correlated facial regions could partially avoid the negative impacts/noises caused by irrelevant facial regions \cite{li2018eac}. Particularly, it first divides the spatio-temporal facial feature $G^t \in \mathbb{R}^{c \times h \times w}$ extracted by MFD module into three subsets corresponding to three slightly overlapped facial regions: (1) the upper region encompassing eyebrows and eyes; (2) the middle region containing the nose and cheeks; and (3) the lower region covering the mouth and chin (Illustrated in Fig. \ref{fig:intro}). This is achieved by directly slicing the feature $G^t$ along the height dimension as:

\begin{equation}
    G^t_{\text{up}}, G^t_{\text{mid}}, G^t_{\text{low}} = G^t[0:\frac{3}{7}h], G^t[\frac{2}{7}h:\frac{5}{7}h], G^t[\frac{4}{7}h:h]
\end{equation}
where the height $h$ of the $G^t$ is $7$ in our implementation, thus we empirically choose this best partition setting. After that, $N$ AU-specific Feature Extractors (\textbf{AFE}) (each is made up of a convolution layer with kernel size of $1 \times 1$ and a Global Average Pooling (GAP) layer) are employed, where the $n_\text{th}$ extractor learns a local relationship-aware feature $v_n^t \in \mathbb{R}^{1 \times b}$ ($b$ denotes the dimension of an AU vector) from its corresponding sliced regional feature ($G^t_{\text{up}}, G^t_{\text{mid}}$ or $G^t_{\text{low}}$), representing the $n_\text{th}$ AU's status at the $t_{th}$ frame. Consequently, each AU feature is extracted in the context of its spatially adjacent AUs (i.e., modelling AU relationship of the same facial region at the feature extraction level).



In addition, the spatio-temporal relationship among AUs of the same facial region are also modelled at their prediction level, where an auxiliary branch (\textbf{Aux}) is added at the training phase. It is trained to predict an AU occurrence combination $Y^t_{\text{sub}} = \{y^t_{\text{sub}, 1}, \cdots, y^t_{\text{sub}, 2^{N_\text{sub}}} \}$ (i.e., $Y^t_{\text{sub}}$ is a one-hot vector and $N_\text{sub}$ is the number of the target AUs in the corresponding sub-region) from each sliced regional feature $G^t_{ \text{sub}} \in \{ G^t_{\text{up}}, G^t_{\text{mid}}, G^t_{\text{low}} \}$, which jointly describes all AUs' occurrence status within each facial region. Mathematically, this process can be formulated as:
\begin{equation}
    P^t_{\text{sub}} = \sigma(\text{FC}_\text{sub}(\text{GAP}(G^t_{ \text{sub}})))
\end{equation}
where $\sigma$ denotes the Softmax function and $\text{FC}_\text{sub}$ denotes a fully connected layer. As a result, training this branch enforces the network encoding underlying local AU relationship to each sliced regional feature, allowing AFE to extract enhanced AU-relevant features from regional features.

\textbf{Cross-regional AU relationship modeling:} Besides spatially adjacent AUs, each AU's activation may also associate with AUs located in other facial regions \cite{li2019semantic}. Consequently, this stage aims to enhance the recognition performance by additionally capturing such cross-regional AU spatio-temporal dependencies within the given face image. It treats each local relationship-aware spatio-temporal AU feature $v_n^t$ extracted in the previous stage as a node, and adaptively connects it with all activated AU nodes belonging to other facial regions (i.e., AU activation status are decided by AU predictions of the first stage). This edge connection definition is inspired by the finding that activated AUs usually have more influences on other AUs \cite{luo2022learning}. As a result, the relationship of each cross regional AU pair is explicitly represented through a graph edge, and further modelled via a Graph Attention Network (GAT) \cite{velivckovic2017graph} layer as:
\begin{equation}
\begin{split}
   & e_{n,m}^t = \text{LeakyReLU} \left (r^{T}\left [ Wv_n^t \parallel Wv_m^t \right ]  \right ) \\
   & \hat{v}_n^t = \phi\left ( \sum_{m\in N_n^t} \alpha_{n,m}^t Wv_m^t  \right )  \\
   & \textbf{Subject to:}  ~~ \alpha_{n,m}^t = \frac{\exp\left ( e_{n,m}^t  \right ) }{ {\sum_{q \in N_n^t} \exp\left ( e_{n,q}^t \right ) } } 
\end{split}
\end{equation}
where $e_{n,m}^t$ is a graph edge defines the impacts of the $m_\text{th}$ AU node to the $n_\text{th}$ AU node in the $t_\text{th}$ frame; $W \in \mathbb{R}^{b \times b}$ denotes a shared linear transformation applied to every node feature; $\parallel$ is the concatenation operation; $\phi$ is an activation function and $r \in \mathbb{R}^{2b}$ denotes the weight of an attention operation.$N_n^t$ is the set of the neighbours of the current node. Subsequently, $N$ local and global hierarchical relationship-aware AU features $\hat{V}^t = \{ \hat{v}_1^t, \hat{v}_2^t, \cdots, \hat{v}_N^t \}$ are generated to describe $N$ target AUs in the $t_\text{th}$ frame.

\begin{table*}[thb]
    \centering
   \scalebox{0.90}{
     \small
     \setlength{\tabcolsep}{1.5mm}{
        \begin{tabular}{clccccccccccccc}
        
        \toprule
        \multicolumn{2}{c}{\multirow{2}{*}{\text{Method}}} & \multicolumn{12}{c}{AU}  & \multirow{2}{*}{\textbf{Avg.}} \\ \cmidrule(lr){3-14}
           & & 1    & 2    & 4    & 6   & 7   & 10   & 12   & 14   & 15   & 17   & 23   & 24   &      \\ \midrule
        \multicolumn{1}{c}{\multirow{4}{*}{\text{\makecell{Static \\image-based}}}}
         & EAC-Net \cite{li2018eac} &39.0  &35.2  &48.6  &76.1   &72.9   &81.9  &86.2   &58.8   &37.5  &59.1  &35.9  &35.8  &55.9  \\
        &JAA-Net  \cite{shao2018deep}         &47.2  &44.0  &54.9  &77.5   &74.6   &84.0  &86.9   &61.9   &43.6  &60.3  &42.7  &41.9  &60.0  \\
       & ARL   \cite{shao2019facial}          &45.8  &39.8  &55.1  &75.7   &77.2   &82.3  &86.6   &58.8   &47.6  &62.1  &47.4  &55.4  &61.1  \\
        &SMA-Net \cite{9666970}               &56.5  &45.1  &57.0  &79.5   &79.5   &84.5  &86.4   &66.1   &55.8  &64.2  &48.7  &56.8  &64.9  \\

        \midrule
        \multicolumn{1}{c}{\multirow{4}{*}{\text{\makecell{Static AU \\relationship modeling}}}}
       & SRERL \cite{li2019semantic}          &46.9  &45.3  &55.6  &77.1   &78.4   &83.5  &87.6   &63.9   &52.2  &63.9  &47.1  &53.3  &62.9  \\
       & FAUDT \cite{jacob2021facial}         &51.7  &49.3  &61.0  &77.8   &79.5   &82.9  &86.3   &67.6   &51.9  &63.0  &43.7  &56.3  &64.2  \\
       & FAN-Trans   \cite{yang2023fan}       &55.4  &46.0  &59.8  &78.7   &77.7   &82.7  &88.6   &64.7   &51.4  &\textbf{65.7}  &50.9  &56.0  &64.8  \\
         &ME-GraphAU \cite{luo2022learning}    &52.7  &44.3  &60.9  &79.9   &80.1   &\underline{85.3}  &\underline{89.2}   &\underline{69.4}   &55.4  &64.4  &49.8  &55.1  &65.5  \\
        \midrule

        \multicolumn{1}{c}{\multirow{6}{*}{\text{Spatio-temporal}}}
        & STRAL  \cite{shao2020spatio}         &48.2  &47.7  &58.1  &75.8   &78.1   &81.6  &87.6   &60.5   &50.2  &64.0  &51.2  &55.2  &63.2  \\
         &HSTR-Net\cite{song2022heterogeneous} &55.5  &49.5  &\textbf{61.9}  &76.6   &80.2   &84.2  &87.4   &62.6   &54.8  &64.1  &47.1  &52.1  &64.7  \\
        &KS \cite{Li_2023_ICCV}               &55.3  &48.6  &57.1  &77.5   &\textbf{81.8}   &83.3  &86.4   &62.8   &52.3  &61.3  &\textbf{51.6}  &\underline{58.3} &64.7  \\
         &WSRTL \cite{yan2022weakly}   &\textbf{59.7}  &\textbf{51.7}  &\underline{61.6}  &\textbf{80.3}   &\underline{80.9}   &85.2  &\textbf{89.7}   &67.8   &52.2  &63.4  &\underline{51.4}  &46.9  &65.9  \\
        \cmidrule(lr){2-15}
       & Ours (ResNet-50)                     &\underline{58.3}  &\underline{50.9}  &58.9  &78.4   &80.3   &84.9  &88.2   &\textbf{69.5}   &\underline{56.0}  &\underline{65.5}  &49.5  &\textbf{59.3}  &\textbf{66.6}  \\
       \multicolumn{1}{l}{} &Ours (Swin-B)                        &54.6 	&49.7 	&61.0 	&\underline{79.9} 	&79.4 	&\textbf{85.4} 	&88.5 	&67.8 	&\textbf{56.8} 	&63.2 	&50.9 	&55.4 	&\underline{66.1}  \\
        \bottomrule
        \end{tabular}}
    }
    \caption{F1 scores (in \%) achieved for 12 AUs on BP4D dataset. The best and the second best results of each column are indicated with bold font and underline, respectively.}
    \label{ex:tab_BP4D_sota}
    \vspace{-0.1cm}
\end{table*}
\begin{table*}[thb]
    \centering
    \scalebox{0.90}{
    \small
    \setlength{\tabcolsep}{1.5mm}{
    \begin{tabular}{clccccccccc}
    \toprule
    \multicolumn{2}{c}{\multirow{2}{*}{\text{Method}}} & \multicolumn{8}{c}{AU}  & \multirow{2}{*}{\textbf{Avg.}} \\ \cmidrule(lr){3-10}
    \multicolumn{2}{l}{}    & 1               & 2       & 4   & 6   & 9   & 12   & 25   & 26     &      \\\midrule
    \multicolumn{1}{c}{\multirow{4}{*}{\text{\makecell{Static \\image-based}}}}
    &EAC-Net \cite{li2018eac}               &41.5   &26.4   &66.4  &50.7   &\textbf{80.5}   &\textbf{89.3}  &88.9   &15.6   &48.5   \\
    &JAA-Net \cite{shao2018deep}            &43.7   &46.2   &56.0  &41.4   &44.7   &69.6  &88.3   &58.4   &56.0   \\
    &ARL \cite{shao2019facial}              &43.9   &42.1   &63.6  &41.8   &40.0   &76.2  &95.2   &66.8   &58.7   \\
    &SMA-Net  \cite{9666970}                &53.4   &54.2   &64.0  &57.0   &47.0   &76.6  &92.0   &55.2   &64.2   \\
    
    \cmidrule(lr){1-11}
    \multicolumn{1}{c}{\multirow{4}{*}{\makecell{Static AU \\relationship modeling}}}
    & SRERL \cite{li2019semantic}            &45.7   &47.8   &59.6  &47.1   &45.6   &73.5  &84.3   &43.6   &55.9   \\
    &FAUDT \cite{jacob2021facial}           &46.1   &48.6   &72.8  &56.7   &50.0   &72.1  &90.8   &55.4   &61.5   \\
    & FAN-Trans \cite{yang2023fan}           &56.4   &50.2   &68.6  &49.2   &57.6   &75.6  &93.6   &58.8   &63.8   \\
    & ME-GraphAU\cite{luo2022learning}       &54.6   &47.1   &72.9  &54.0   &55.7   &76.7  &91.1   &53.0   &63.1   \\
    
    \midrule
    \multicolumn{1}{c}{\multirow{6}{*}{\text{Spatio-temporal}}}
    &STRAL\cite{shao2020spatio}             &52.2   &47.4   &68.9  &47.8   &56.7   &72.5  &91.3   &\underline{67.6}   &63.0  \\
    &HSTR-Net\cite{song2022heterogeneous}   &54.3   &50.8   &70.1  &\textbf{66.6}   &\underline{59.6}   &68.0  &\textbf{97.9}   &\textbf{69.8}   &62.9  \\      
    &KS \cite{Li_2023_ICCV}                 &53.8   &\underline{59.9}   &69.2  &54.2   &50.8   &75.8  &92.2   &46.8   &62.8  \\
    & WSRTL \cite{yan2022weakly}             &57.3   &51.8   &\underline{74.3}  &49.8   &44.8   &\underline{79.3}  &94.6   &\underline{64.6}   &64.6  \\
    
   \cmidrule(lr){2-11}
    &Ours (ResNet-50)                       &\underline{61.4}   &57.7   &70.9  &\underline{57.1}   &48.3   &75.7  &91.5 	 &56.7 	 &\underline{64.9} \\
    &Ours (Swin-B)                          &\textbf{65.4}   &\textbf{60.2}   &\textbf{75.2}  &50.2   &52.4   &74.3  &93.7   &58.2   &\textbf{66.2} \\
    \bottomrule
    \end{tabular}
    }}
    \caption{F1 scores (in \%) achieved for 8 AUs on DISFA dataset. The best and second best results of each column are indicated with bold font and underline, respectively.}
    \label{ex:tab_DISFA_sota}
    \vspace{-0.4cm}

\end{table*}

\subsection{Loss function}

As AU recognition constitutes a multi-label binary classification task, with most AUs inactivated the across majority of frames (please refer to Supplementary Material for details), an asymmetric loss function \cite{luo2022learning} is employed. Given the input consecutive $T$ facial frames with $N$ target AUs, the loss function $\mathcal{L}_\text{AU}$ for supervising all AUs' recognition (i.e., output by the TCN/SC layers) is defined as:
\vspace{-0.3cm}
\begin{equation}
\label{asymmetric loss}
\mathcal{L}_\text{AU} = - \sum_{n=1}^{N} \sum_{t=1}^{T} w_n [y_{n}^{t} \log(p_{n}^{t}) + p_{n}^{t} (1-y_{n}^{t})  \log(1-p_{n}^{t})]
\end{equation}
where $p_{n}^{t}$ and $y_{n}^{t}$ are the $n_\text{th}$ AU's prediction and the corresponding ground-truth for the frame $f^t$, respectively; a $w_n$ is calculated for each AU based on the training set to alleviate label imbalance issue; the $p_{n}^{t}$ at the beginning of the second term $p_{n}^{t} (1-y_{n}^{t}) \log(1-p_{n}^{t})$ dynamically down-weights the contribution of negative samples (inactive AUs), as inactive AUs significantly outnumber active ones in the training set. 
Besides, a cross-entropy loss is utilized to individually supervise regional AU combination predictions of all frames produced by the first stage of the HSR module as:
\begin{equation}
\mathcal{L}_\text{sub} =\sum_{t=1}^{T} \sum_{\text{sub} = \{\text{up,mid,down}\}}{}\text{CE}(P^t_{\text{sub}},Y^t_{\text{sub}})
\end{equation}
where $P^t_\text{sub}$ and $Y^t_{\text{sub}}$ denote the AU combination prediction and the corresponding ground-truth of a facial region in the frame $f^t$ and $\text{CE}$ denotes the cross-entropy function. By predicting such AU combinations consisting of multiple AUs located in the same facial region, the network is encouraged to model underlying dependencies among spatially adjacent AUs in each facial region. Consequently, the overall loss function for training the proposed network combines the two loss functions described above as:
\begin{equation}
\label{loss_weight}
    \mathcal{L} = \mathcal{L}_\text{AU} +\lambda \mathcal{L}_\text{sub}
\end{equation}
where $\lambda$ balances the contribution of the two losses. 



\section{Experiments}

\subsection{Experimental setup}

\noindent \textbf{Datasets:} Our MDHR is evaluated on two AU recognition benchmark datasets: BP4D \cite{zhang2014bp4d} and DISFA \cite{mavadati2013disfa}. BP4D is made up of 328 facial videos containing around 140,000 frames collected from 23 females and 18 males. Meanwhile, DISFA contains 27 facial image sequence (totally $130,815$ frames) recorded from 12 females and 15 males who were asked to watch Youtube videos. Each frame in BP4D and DISFA is annotated with occurrence labels corresponding to 12 and 8 AUs, respectively.

\noindent \textbf{Implementation details:} We follow previous approaches \cite{Zhang_2018_CVPR,shao2021jaa} to apply MTCNN \cite{zhang2016joint} to crop and align a $224 \times 224$ face region from each frame, and conduct subject-independent three-folds cross-validation for each dataset, where the reported results are achieved by averaging the validation results of three folds. We pad $k$ frames that same to the first frame / last frame at the beginning / end of each face video to ensure all frames can be processed by our model. AdamW \cite{loshchilov2018decoupled} optimizer with $\beta_1 = 0.9$, $\beta_2 = 0.999$ is employed for training and the $\lambda$ in Eq. \ref{loss_weight} is set to 0.01. A cosine decay learning rate scheduler is utilized, with an initial value of $0.0001$. Both backbones are pre-trained on ImageNet \cite{deng2009imagenet}. All our experiments are conducted using NVIDIA A100 GPUs based on the open-source PyTorch library. More detailed model, training/validation, and dataset settings are provided in the Supplementary Material.

\noindent \textbf{Metrics:} Following previous AU recognition studies \cite{shao2021jaa,churamani2021aula,li2019self}, a common metric: frame-based F1 score ($F1 = 2 \frac{P \cdot R}{P+R}$), is employed to evaluate the performance of our MDHR, which takes both recognition precision $P$ and recall rate $R$ into consideration.

\subsection{Comparison with state-of-the-arts}

Table \ref{ex:tab_BP4D_sota} and Table \ref{ex:tab_DISFA_sota} compare our approach with previous state-of-the-art AU recognition methods, including eight static image-based methods \cite{li2018eac,shao2018deep,shao2019facial,9666970,li2019semantic,jacob2021facial,yang2023fan,luo2022learning} (where four methods specifically conduct AU relationship modeling) and four spatio-temporal methods \cite{shao2020spatio,song2022heterogeneous,Li_2023_ICCV,yan2022weakly}. It can be observed that our MDHR achieved new SOTA results on both datasets, with F1-scores of 66.6\% (ResNet50 backbone) and 66.2\% (Swin-Transformer backbone) on BP4D and DISFA, respectively. Particularly, it has clear advantages over all static image-based methods, e.g., outperformed previous state-of-the-art static AU relationship modelling method \cite{luo2022learning} with 1.1\% (ResNet) and 0.6\% (Swin-Transformer) improvements on BP4D, as well as 1.8\% (ResNet) and 3.1\% (Swin-Transformer) improvements on DISFA, respectively. Meanwhile, our approach is also superior to previous spatio-temporal methods, achieving 0.7\% and 1.6\% higher F1 results over the best model WSRTL \cite{yan2022weakly} on BP4D and DISFA, respectively.

These results suggest that: (i) the proposed MDHR is effective and robust in AU recognition, as it achieved both best and the second best performances on both datasets under two backbone settings; (ii) jointly modelling spatio-temporal relationship among AUs could lead to additional performance gains compared to approaches \cite{luo2022learning,song2021uncertain} that only consider their spatial relationship; and (iii) our MDHR can better capture AU-related spatio-temporal cues over existing spatio-temporal AU recognition approaches \cite{yan2022weakly,song2022heterogeneous}. We didn't compare approaches that utilized additional face datasets to train AU models \cite{yang2023toward} despite our MDHR still clearly outperformed them.

\begin{table}[t]
    \centering
    \small
    \scalebox{0.9}{
    \setlength{\tabcolsep}{2.0mm}{
    \begin{tabular}{cccccc|c}
    \toprule
     \multirow{2}{*}{\text{Backbone}} &   \multirow{2}{*}{\text{MFD}} & \multicolumn{3}{c}{\text{HSR}} & \multicolumn{1}{c|}{\multirow{2}{*}{\text{TCN}}} &\multirow{2}{*}{\text{F1-score}} \\ \cline{3-5}
     & & \multicolumn{1}{c}{AFE}  & Aux & CRM &  & \\
    \midrule
     \checkmark &        &        &        &        &     &   63.3 \\
     \checkmark & \checkmark &    &        &        &      &   64.6  \\
     \checkmark &             & \checkmark    &        &        &      & 64.1\\
    \checkmark &  &\checkmark &   \checkmark     &        &    &  64.5  \\
    \checkmark &  &  \checkmark    &   \checkmark      &   \checkmark     &     &  65.1 \\
     \checkmark & \checkmark &\checkmark   &   &     &       &   65.3  \\
     \checkmark & \checkmark &\checkmark & \checkmark  &      &       &   65.7  \\
     \checkmark & \checkmark & \checkmark   & \checkmark  & \checkmark &  &    66.3     \\
     \checkmark & \checkmark &\checkmark & \checkmark  & \checkmark & \checkmark &   \textbf{66.6} \\
     \bottomrule
    \end{tabular}
    }}
    \caption{Average AU recognition F1 scores (\%) achieved by various settings using ResNet50 backbone on BP4D dataset, where \textbf{AFE}, \textbf{Aux}, and \textbf{CRM} denote the AU-specific feature extractor, the added auxiliary branch and the Cross-regional AU relationship modelling block, all of which belong to the HSR module.}
    \label{ex:tab_AblationStudy}
    \vspace{-0.3cm}

\end{table}

\subsection{Ablation studies}

We perform ablation studies on BP4D dataset to demonstrate various aspects of our approach, where the default setting employs the ResNet as the backbone and asymmetric loss (Eqa. \ref{asymmetric loss}) for the model training. We further provide more ablation results (e.g., the influence of the number of adjacent frames $k$, statistical analysis, model complexity analysis, etc.) in the Supplementary Material.

\textbf{Contribution of each component:} Table \ref{ex:tab_AblationStudy} compares contributions of different modules. Firstly, our MFD module brought 1.3\% absolute improvement, highlighting the effectiveness of the MFD in capture AU-related spatio-temporal facial behaviour cues. Meanwhile, individually employing the HSR module boosting the F1 score from 63.3\% to 65.1\%, validating the importance of modelling hierarchical spatio-temporal relationship among AUs. Specifically, the use of AU-specific feature extractors to individually learn each AU from its sliced facial region improved the F1 score from 63.3\% to 64.1\%, and the auxiliary branch also contributes additional 0.4\% improvement. Finally, we found that combining our MFD and HSR module with the TCN resulted in the best performance, which validates that these two modules can learn complementary AU-related cues to further enhance AU recognition performance.





\begin{figure}
    \centering
    \includegraphics[width=0.96\linewidth]{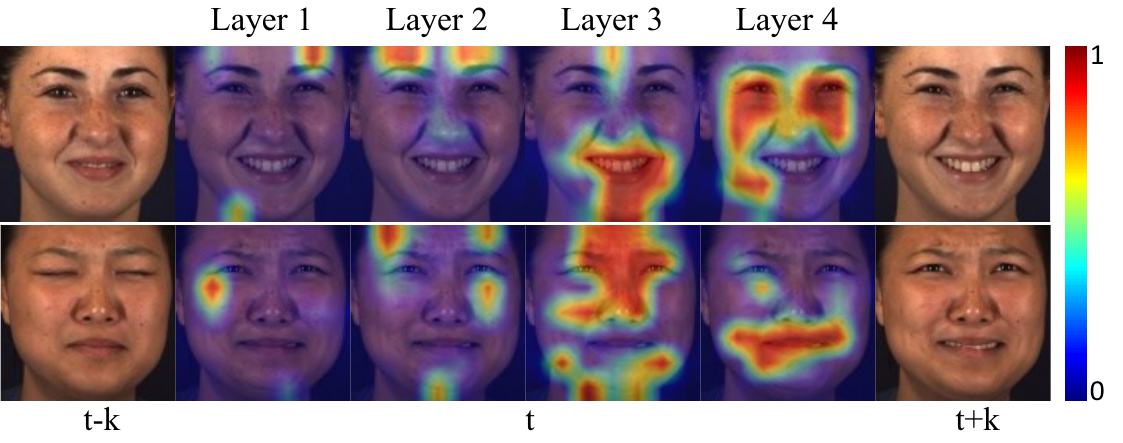}
    \caption{Visualization of adaptive weight matrices learned by the MFD module. The weight matrices learned for feature maps of shallow layers (layer 1 and 2) emphasized subtle motions (e.g., subtle eyebrow and cheek motions), while large check and mouth movements are captured in deeper layers (layer 3 and 4).}
    \label{fig:vis_weight}
    \vspace{-0.3cm}
\end{figure}

\begin{table}[ht]
    \centering
    \scalebox{0.9}{
    \setlength{\tabcolsep}{0.85mm}{
    \begin{tabular}{c|c|cc|ccc|ccc}
    \toprule
         Method  & baseline& \multicolumn{2}{c}{\text{Alone}} & \multicolumn{3}{|c|}{\text{Combination}} & Sum & Cat & AW  \\
    \midrule
         layer 1 &           & \checkmark  &            & \checkmark &           & \checkmark &\checkmark&\checkmark&\checkmark \\
         layer 2 &           &             &           & \checkmark &            & \checkmark &\checkmark&\checkmark&\checkmark \\
         layer 3 &           &             &            &            & \checkmark & \checkmark &\checkmark&\checkmark&\checkmark  \\
         layer 4 &           &             & \checkmark &            & \checkmark &             &\checkmark&\checkmark&\checkmark \\
    \midrule
        F1-score& 64.1&64.3&64.7&64.6&64.8&64.7&64.9&64.9&\textbf{65.3} \\
    \bottomrule
    \end{tabular}
    }
    }
    \caption{Results of different MFD module settings, where the left part displays results achieved by computing facial dynamic at different layers and their combinations (combined using our adaptive weighting), while the right side displays results of two other fusion strategy applied to combine facial dynamic of all scales, and AW denotes adaptive weighting.}
    \label{tab:effectiveness of adaptive weighting}
    \vspace{-0.5cm}
\end{table}

\textbf{Analysis of the MFD module:} Table \ref{tab:effectiveness of adaptive weighting} investigates our MFD module based on the system (baseline) that combines the backbone and AU-specific feature extractors. It is clear that even using facial dynamic learned from the outputs of a single and two backbone layers can consistently benefit the recognition, while combining dynamic of all scales resulted in the largest improvement. This suggests that facial dynamic extracted by our MFD at different spatial scales contain complementary and useful cues for AU recognition., i.e., our MFD can emphasize each AU-related cues at its most related spatial scales (illustrated in Figure \ref{fig:vis_weight}). Although simply adding or concatenating unweighted dynamic features of all spatial scales can already lead to performance gains, our adaptive weighting block still show clear advantage over them, suggesting that it can effectively consider the importance of each spatial scale on different AUs' recognition.



\begin{table}[ht]
    \centering
    \scalebox{0.9}{
    \begin{tabular}{cc}
    \toprule
       AU relationship modeling method  & F1-score    \\
    \midrule
         Fully-connected                                     &   64.6   \\ 
         Aux + Locally connected                             &   64.2 \\
         Aux + Cross-regional fully-connected                &   64.9   \\
         Aux + Fully connected                               &   64.8  \\
         Aux + Connecting cross-regional activated AUs       &  \textbf{65.1}   \\
    \bottomrule
    \end{tabular}
    }
    \caption{Results of different edge connection strategies.}
    \label{tab:effectiveness of hierarchical relationship}
    \vspace{-0.3cm}
\end{table}

\begin{figure}
    \centering
    \includegraphics[width=1\linewidth]{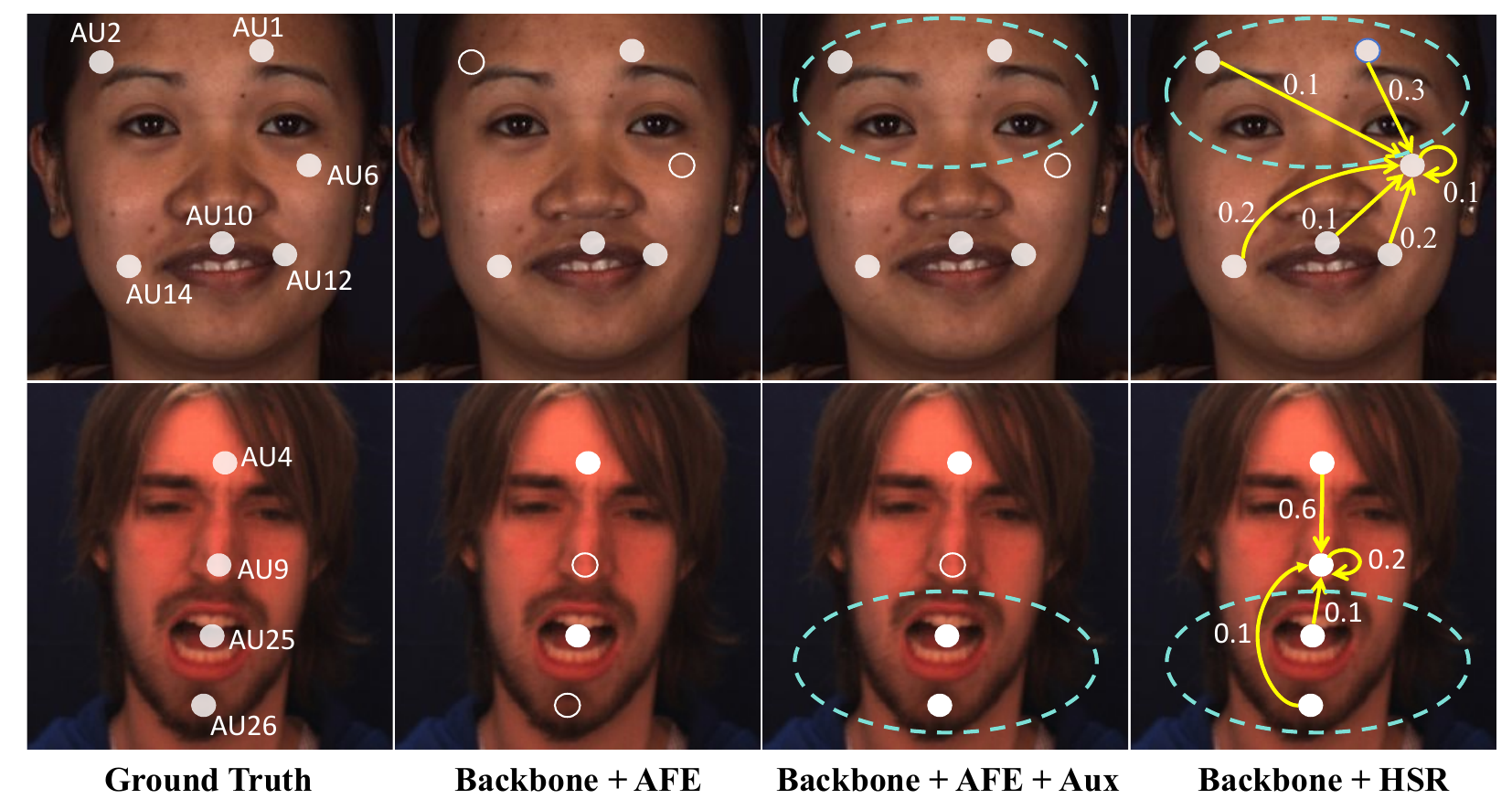}
    \caption{Visualization of AU predictions under three HSR settings, where white solid and hollow dots denote activated and inactivated AUs. The green dotted circles denote the local AU relationship modelling, while the yellow lines/weights denote the graph edges describing the association between AUs. It can be observed that the local relationship modelling can effectively model dependencies between AUs in the same region to make better predictions (e.g., AU2 and AU26 in column 3), while additionally use cross-regional AU relationship modelling can further utilize the learned relationship cues to improve AU predictions in different facial regions (e.g., AU6 and AU9 in column 4). }
    \label{fig:vis_graph}
    \vspace{-0.4cm}
\end{figure}

\textbf{Analysis of the HSR module:} Table \ref{ex:tab_AblationStudy} first demonstrates that not only the HSR module brought clear improvements but also both of its local and cross-regional AU relationship modelling blocks can improve AU predictions, i.e., all its block (e.g., AFE, Aux and CRM) positively contributed to the final performance. Figure \ref{fig:vis_graph} further visualizes the impact of these blocks. Additionally, Table \ref{tab:effectiveness of hierarchical relationship} compares different AU graph edge connection settings of the cross-regional AU relationship modelling block, where the setting that connects each activated AU to all other AUs located in different facial region achieved the superior performance to other edge settings, i.e., this setting can effectively model cross-regional AU relationship. Importantly, the HSR is not sensitive to different edge connection settings when cross-regional AU relationship is considered.

\section{Conclusion}

This paper proposes a novel MDHR that not only computes facial dynamics at different spatial scales as AUs could exhibit heterogeneity in their ranges and magnitudes, but also models hierarchical spatio-temporal relationships among AUs. Results show that the proposed two modules can effective capture AU-related dynamics and their relationships, making our MDHR becoming the new SOTA AU recognition method. The main limitations are that our facial region slicing strategy could be potentially improved and more advanced graph edge learning strategies could be applied to HSR for better modelling relationships.

\section{Acknowledgement}

The work is supported by National Natural Science Foundation of China under Grant 82261138629; Guangdong Basic and Applied Basic Research Foundation under Grant 2023A1515010688 and Shenzhen Municipal Science and Technology Innovation Council under Grant JCYJ20220531101412030.

{
    \small
    \bibliographystyle{ieeenat_fullname}
    \bibliography{main}
}


\end{document}